\documentclass[letterpaper,10pt, conference]{IEEEtran} 
\IEEEoverridecommandlockouts                         
\usepackage[numbers]{natbib}
\usepackage{amsmath}
\usepackage{mathabx}
\usepackage[bookmarks=true]{hyperref}
\usepackage{graphicx}
\usepackage{times}
\usepackage{underscore}
\usepackage{moreverb,url}
\usepackage[dvipsnames]{xcolor}
\usepackage{url}
\usepackage{times}
\usepackage{multicol}
\usepackage[utf8]{inputenc}
\usepackage{graphics}
\usepackage{bm}
\usepackage[linesnumbered,ruled,noend]{algorithm2e}
\usepackage[noend]{algpseudocode}
\usepackage{multicol}
\usepackage{lipsum}
\usepackage[english]{babel}
\usepackage{blindtext}
\usepackage{mathtools}
\usepackage{booktabs}

\renewcommand\vec{\mathbf}

\SetKwRepeat{Do}{do}{while}
\DeclareMathOperator*{\argmax}{arg\,max}

\usepackage{etoolbox}
\makeatletter
\patchcmd{\@makecaption}
  {\scshape}
  {}
  {}
  {}
\makeatletter
\patchcmd{\@makecaption}
  {\\}
  {.\ }
  {}
  {}
\makeatother
\def\tablename{Table}

\makeatother
\title{\vspace{5mm}Learning to Efficiently Plan Robust Frictional Multi-Object Grasps}

\author{Wisdom C. Agboh$^{*1,2}$, Satvik Sharma$^{*1}$, Kishore Srinivas$^{1}$, Mallika Parulekar$^{1}$ \\ Gaurav Datta$^{1}$, Tianshuang Qiu$^{1}$, Jeffrey Ichnowski$^{3}$, Eugen Solowjow$^{4}$,
Mehmet Dogar$^{2}$, Ken Goldberg$^{1}$
\thanks{$^{1}$University of California, Berkeley, USA.}%
\thanks{$^{2}$University of Leeds, UK.}%
\thanks{$^{3}$Carnegie Mellon University, Pittsburgh, USA.}%
\thanks{$^{4}$Siemens Research Lab, Berkeley, USA.}%
\thanks{$^{*}$equal contribution}%
}

\begin{document}

\maketitle
\begin{abstract}
We consider a decluttering problem where multiple rigid convex polygonal objects rest in randomly placed positions and orientations on a planar surface and must be efficiently transported to a packing box using both single and multi-object grasps. Prior work considered frictionless multi-object grasping. 
\textcolor{black}{In this paper, we introduce friction to increase the number of potential grasps for a given group of objects, and thus increase picks per hour.}
We train a neural network using real examples to plan robust multi-object grasps.
In physical experiments, we find a 13.7\% increase in success rate, a 1.6x increase in picks per hour, and a 6.3x decrease in grasp planning time compared to prior work on multi-object grasping. Compared to single-object grasping, we find a 3.1x increase in picks per hour. 

\end{abstract}

\section{Introduction}
\begin{figure}[t]
    \centering
    \includegraphics[scale=0.29]{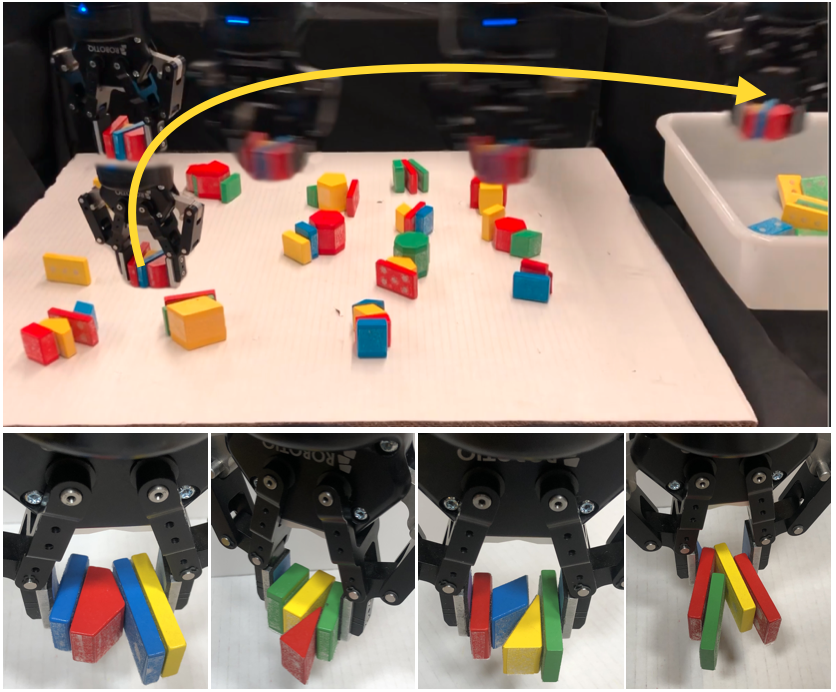}
    \caption{ The decluttering problem (top) where objects must be transported to a packing box. We find robust frictional multi-object grasps (bottom) to efficiently declutter the scene.}
    \label{fig:fig1}
    \vspace{-5mm}
\end{figure}

When skilled waiters clear tables, they grasp multiple utensils and dishes in a single motion. Similarly, it is inefficient for robotic picking systems in warehouses and fulfillment centers to only handle a single object at a time. Picking multiple objects at once can significantly increase picks per hour (PPH), the total number of objects picked from a scene in an hour. In prior work on multi-object grasping \cite{Agboh-ISRR-2022}, PPH was increased compared to single-object picking. This improvement was limited due to a frictionless grasping assumption and no considerations of robustness. In this work, we find that considering friction and quickly generating robust grasps can lead to significant improvements in PPH.
For example, grasps like those shown in Fig.~\ref{fig:fig1} cannot exist without appropriate friction between objects. An important question that then arises is how to generate such robust frictional grasps.

Robust grasps have been generated in prior work~\cite{Mahler-RSS-2017, Shen-RSS-22, Liang-Access-2021, jiang2020textureless}, but only for single objects. Inspired by these works, we develop a robust multi-object grasping system for planar convex polygonal objects. Instead of using a physics simulator, we propose to collect data entirely on a physical robot and use it to train a multi-object grasping function, \textit{MOG-Net}, which is robust to state and control uncertainty and predicts the number of objects that will be grasped out of a target object group. We train in real to avoid the sim-to-real gap \cite{Dongwon-IROS-2020, Ding-IROS-2021, Chebotar-ICRA-2019}. We also propose a necessary condition for frictional multi-object grasping to filter out inadmissible grasps and show that this filtering leads to a high quality dataset and saves valuable physical robot time during data collection.

We use \textit{MOG-Net} in a novel grasp planner to generate robust multi-object grasps. The planner maximizes the predicted number of objects grasped per pick attempt in cluttered scenes. To improve robustness to state and control uncertainty, we weight \textit{MOG-Net's} predictions with the probability of satisfying multi-object grasping necessary conditions, obtained via Monte-Carlo sampling.

We find a 13.7\% increase in success rate, a 1.6x increase in picks per hour, and an 6.3x decrease in grasp planning time compared to prior work~\cite{Agboh-ISRR-2022} on multi-object grasping. Compared to single-object grasping we find a 3.1x increase in picks per hour. This work makes 4 contributions:

\begin{enumerate}
    \item The derivation of a frictional multi-object grasping necessary condition to filter inadmissible grasps.
    \item \textit{MOG-Net}: a robust multi-object grasp neural network, self-supervised in real to predict the number of objects grasped out of a target group, given a grasp candidate.
    \item A grasp planning algorithm, $\mu$-MOG, that generates grasps that are robust to state and control uncertainty, by considering the probability of necessary conditions being satisfied.
    \item Physical experiments evaluating the methods with randomized scenes of 58 polyhedral objects totaling 2532 grasp attempts.
\end{enumerate}

\section{Related work}

Decluttering or picking multiple objects from a table is a common robotics problem \cite{mahler2017binpicking} which has mainly been addressed with single object grasps \cite{Morrison-IJRR-2020, Lou-ICRA-2021}. In this section, we discuss prior work on multi-object grasping, frictional single object grasping, and robust grasp synthesis.
 
\subsection{Multi-object grasps}
\citet{Harada-ICRA-1998} proposed some of the earliest works on multi-object grasping. In ~\cite{Harada-ICRA-1998, Harada-IROS-1998, Harada-TRA-2000} they develop conditions for enveloping grasps of multiple objects using a multi-fingered robot hand, and under a rolling contact assumption. Not long after, \citet{Yamada-ICRA-2005} proposed a series of methods~\cite{Yamada-ICRA-2005, Yamada-ISMNHS-2005,Yamada-ROBIO-2009,  Yamada-ICMA-2012, Yamada-JCSE-2015} to evaluate the grasp stability of multiple planar objects grasped by a multi-fingered robot hand. While these were the pioneering works on multi-object grasping, their results focused on numerical simulations, without physical robot multi-object grasps. This paper derives conditions for equilibrium multi-object grasps, under the frictional point-contact model and shows physical robot multi-object grasps.

Recently, \citet{Chen-IROS-2021} investigated the problem of dipping a robot hand inside a pile of identical spherical objects, closing the hand and estimating the number of objects remaining in the hand after lifting. \citet{Shenoy-CoRR-2021} focused on the same problem but with a goal of transferring the picked spherical objects to another bin. Our work is focused on frictional multi-object grasps of arbitrary convex polygonal objects spread over a plane, so physics-based planning is required~\cite{Dogar-RSS-2011, Danielczuk-CASE-2018, Huang-IROS-2021, Agboh-Humanoids-2018}.

\citet{Sakamoto-IROS-2021} proposed a picking system that first uses robot pushing~\cite{largescalerearrange, Bejjani-IROS-2021, Mohammed-ICRA-2020}, to move one cuboid to the other and thereafter grasps both cuboids in \textcolor{black}{separate actions}. In this paper, we take a single-step push-grasp action and derive conditions for multi-object grasping under the frictional point contact model. Given the uncertainty in grasping systems~\cite{Agboh-WAFR-2018, Agboh-arxiv-2021}, we plan robust multi-object grasps to improve PPH.

\subsection{Frictional grasps}
Some prior works on single object grasping have studied the effect of friction. \citet{golan2020grasps} develop a gripper that can switch between being frictionless and frictional, and show that having a frictional gripper provides more secure and robust grasps in real. \citet{hang2013graspsynthesis} propose a measure of friction sensitivity to assess how grasp quality varies with the coefficient of friction. Inspired by nature, \citet{Roberge-RAL-2018} use a gecko-inspired adhesive on a gripper to apply large shear forces with low normal forces for grasping a single object. \citet{Agboh-ISRR-2022} did not consider friction in multi-object grasping. However, for single objects, friction can increase the number of stable grasps. Using a point contact model, the friction cone is larger for a higher coefficient of friction, meaning that more stable single object grasps can be found. We show that the same is true for multi-object grasping. We focus on frictional multi-object grasps where objects are pushed together before they are grasped.

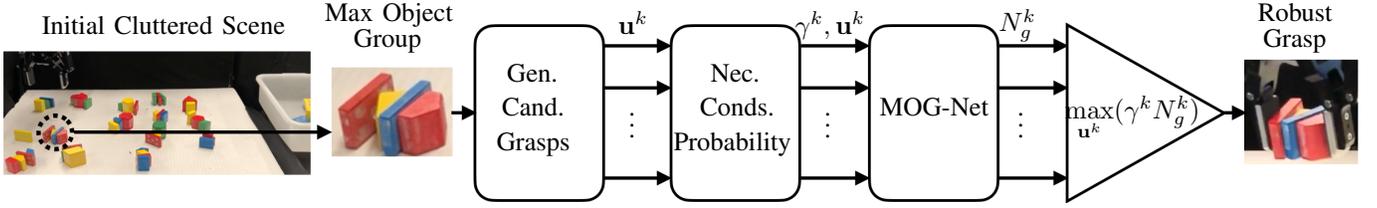
\begin{figure*}[htb!]
    \centering
    \vspace{5mm}
    \includegraphics[scale=0.208]{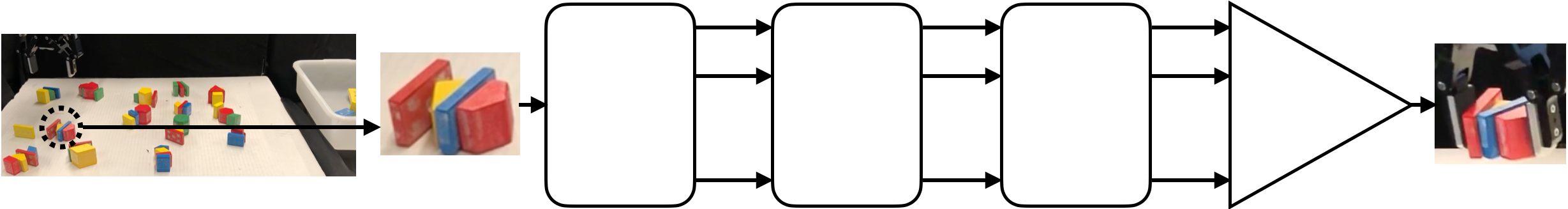}
    \begin{picture}(0,0)
    \put(-240, 76) {Initial Cluttered Scene}
    \put(-133, 81) {Max Object}
    \put(-122, 71) {Group}
    \put(-64, 58) {Gen.}
    \put(-66, 45) {Cand.}
    \put(-68, 32) {Grasps}
     \put(13, 58) {Nec.}
    \put(9, 45) {Conds.}
    \put(-1, 32) {Probability}
    \put(77,45){MOG-Net}
    \put(122,77){$N^{k}_{g}$}
    \put(45,76){$\gamma^{k},\vec{u}^{k}$}
    \put(-22,76){$\vec{u}^{k}$}
    \put(220, 81){Robust}
    \put(222,71){Grasp}
\put(147, 44) {$\max \limits_{\vec{u}^{k}}(\gamma^{k} N^{k}_{g})$}
    \put(129,37){\vdots}
    \put(56, 37){\vdots}
    \put(-18, 37){\vdots}
    \end{picture}
    \vspace{-3mm}
    \caption{An overview of the decluttering system proposed in this paper. It finds the maximum group of objects that can fit in the gripper and generates a robust grasp for that group. First, it generates candidate grasps, and for each grasp $\vec{u}^{k}$, estimates a probability of satisfying multi-object grasp necessary conditions ($\gamma^{k}$), under state and control uncertainty. Thereafter, it uses \textit{MOG-Net} which was trained in real to predict the number of objects ($N^{k}_{g}$) that will be grasped using $\vec{u}^{k}$. The chosen robust grasp maximizes the product $\gamma^{k} \cdot N^{k}_{g}$. We execute the robust grasp and continue to the next object group until the table is cleared of all objects.}
    \label{fig:overview}
    \vspace{-4mm}
\end{figure*}

\subsection{Robust grasping}
Uncertainty in state and model disparity result in grasp failure. An important line of grasping work focuses on generating grasps that are robust to uncertainty. To the best of our knowledge there has been no prior work on robust multi-object grasping. Prior work on robust single object grasping typically falls into one of two categories --- analytic or data-driven.

One analytic approach to robust grasp synthesis is the use of funnels \cite{Agboh-arxiv-2021, Majumdar-IJRR-2017}. These are primitives that can analytically reduce or `funnel' uncertainty from some initial set to a smaller target set. \citet{Bhatt-RSS-21} use a sequence of funnels to perform open-loop and robust in-hand manipulation. Another analytic method is caging \cite{Diankov-Humanoids-2008, Makita-AR-2017, Rodriguez-IJRR-2012} where a grasped object's mobility is bounded such that it `follows' when the gripper moves. These analytic methods exploit the object's environment and shape, using carefully chosen primitives to generate robust grasps. Our work focuses on data-driven methods to generate robust grasps.

Popular data-driven single-object robust grasp synthesis approaches are Dex-Net 2.0 and Dex-Net 4.0 \cite{Mahler-RSS-2017} \cite{dexnet4.0}. They train a grasp quality convolutional neural network (GQ-CNN) with synthetic data to predict grasp success probability. Similarly, Dex-Net 3.0 trains a GQ-CNN for robust grasps on single objects but for suction cups \cite{dexnet3.0}.

There is a sim-to-real gap \cite{Dongwon-IROS-2020, Ding-IROS-2021, Chebotar-ICRA-2019} when grasps are trained in simulation. Robustness in these simulation settings has been achieved through domain randomization or general Monte-Carlo sampling \cite{Dong-TME-2022, Satish-RAL-2019, Tobin-IROS-2018}.

In this work, we propose a frictional multi-object grasp necessary condition and use it to filter inadmissible multi-object grasp candidates. To be robust to state and control uncertainty, we estimate the probability of satisfying these necessary conditions through Monte-Carlo sampling. We avoid the sim-to-real problem by training \textit{MOG-Net} entirely in real.

\section{Problem Statement}

We consider a decluttering problem where multiple rigid convex polygonal objects rest in randomly placed positions and orientations on a planar surface, visible from an overhead camera, and must be transported to a packing box. The objective of this work is to develop a decluttering algorithm that maximizes picks per hour (PPH) for this problem, using robust frictional grasps. Note that we do not consider rearrangement actions (e.g. pushing actions) for arranging the groups before multi-object grasping in this work. Finding the optimal rearrangement plan in such scenes is a challenging long-horizon problem that deserves a separate thorough treatment.

\subsection{Assumptions}\label{sec:assumptions}
We assume that the gripper is a parallel-jaw gripper. \textcolor{black}{We acknowledge that multi-fingered grippers provide more opportunities for multi-object grasps but we focus on parallel-jaw grippers as they are common.} We assume
that objects are extruded convex polygons \textcolor{black}{laying on a flat, uniform color surface}, and that we have a set value for the lower bound for $\mu$, the coefficient of \textcolor{black}{(Coloumb)} friction, for all contact interactions. We also assume antipodal multi-object grasps where each object is kept
in equilibrium by two neighbouring objects, or one object and a gripper jaw. \textcolor{black}{ We further assume that a group of objects in force closure will be securely grasped during motion, and neither the grasping force nor the speed of the motion will dislodge the objects.}

\setlength{\textfloatsep}{2mm}
\begin{algorithm}[b]
    \SetKwInOut{Input}{Input}
    \SetKwInOut{Output}{Output}
    \SetKwInOut{Parameters}{Parameters}
    \SetKwInOut{Subroutines}{Subroutines}
    \Do{$N_{o}$ $>$ $\mathrm{0}$ $\mathbf{and}$ $\mathrm{time}$ $\mathrm{remaining}$}{
    $\vec{x}$, $N_{o}$ $\gets$ GetCurrentState(.)\\
    obj_groups $\gets$ CreateObjGroups($\vec{x}$)\\
    ranked_obj_groups $\gets$ RankObjGroups(obj_groups) \\
    \For{obj_group $\mathbf{in}$ $\mathrm{ranked{\_}obj{\_}groups}$}{
    $\vec{u}$ $\gets$ RobustGraspPlanner($\vec{x}$, \textit{obj_group}) \\

    \If {$\vec{u} \neq \{\}$}{
     Execute $\vec{u}$\\
     break 
     }}
     }
    \caption{Decluttering Algorithm}\label{alg:decluttering-algorithm}
\end{algorithm}
\setlength{\floatsep}{2mm}

\subsection{State and action}
The state $\vec{x}$ is a list of all convex polygonal objects, where each object, $\vec{o_{i}}$, in the list is represented by its vertices:
$$\vec{x} =	[ \, {\vec{o_{i}} \, ] \hspace{2mm} \forall i \in [0 \dots N_{o} - 1]}$$
$$\text{where} \hspace{2 mm} \vec{o_{i}} = [\{x^{0}_{i}, y^{0}_{i}\}, \dots, \{x^{v}_{i}, y^{v}_{i}\}, \dots,\{x^{N_{v}-1}_{i}, y^{N_{v}-1}_{i}\} ]$$
Here, $N_{o}$ is the number of objects on the table, $\{x^{v}_i$, $y^{v}_i\}$ represents the 2D position of vertex $v$ of object $i$, provided by an overhead camera, and $N_{v}$ is the maximum number of vertices for each object $i$.

We represent single and multi-object grasp actions in the same way:
    $\vec{u} = [x_{g}, y_{g}, \theta_{g}]$
, where $x_{g}, y_{g}$, and $\theta_{g}$ represent the desired grasp pose of the gripper, after which the jaws close with a maximum force $f_{g}$.

\section{Decluttering with multi-object grasps}

We present an overview of the decluttering system in Fig.~\ref{fig:overview}. Given an initial cluttered scene, we take a greedy approach and find an object group with the maximum number of objects that can fit in the gripper. The next step is to plan a robust multi-object grasp for this object group. We sample candidate grasps within the convex hull of the objects (see Sec.~\ref{sec:rob-mog-plan}).  Thereafter, we estimate the probability $\gamma^{k}$ that the kth grasp $\vec{u}^{k}$ will satisfy the multi-object grasp necessary conditions (see Sec.~\ref{sec:frictional-mog}), under state and control uncertainty. We also query \textit{MOG-Net} (see Sec.~\ref{sec:mog-net}) to predict the number of objects, $N_{g}^{k}$ that the given grasp will successfully pick. Finally, we choose the grasp that maximizes the robust prediction $\gamma^{k} \cdot N_{g}^{k}$.  In the following subsections, we provide details of the decluttering algorithm and the robust multi-object grasp planner.

\subsection{Decluttering}

\setlength{\textfloatsep}{2mm}
\begin{algorithm}[b]
    \SetKwInOut{Input}{Input}
    \SetKwInOut{Output}{Output}
    \SetKwInOut{Parameters}{Parameters}
    \SetKwInOut{Subroutines}{Subroutines}
    \Input{$\vec{x}$: Current state \\
    obj_group: Objects in the potential grasp}
    \Output{$\vec{u}^{r}$: A robust grasp action \\ $N_{g}^{r}$: Predicted number of objects with $\vec{u}^{r}$}

    grasp_cands $\gets$ GenGraspCands($\vec{x}$, obj_group) \\
    \For{$\vec{u}^{k}$ $\mathbf{in}$ $\mathrm{grasp\_cands}$}{
        $\gamma^{k}$ $\gets$ NecessaryCondsProba($\vec{x}$, $\vec{u}^{k}$) \\
        $N^{k}_{g}$ $\gets$ MOG-Net($\vec{x}$, $\vec{u}^{k}$) \\
    }
    $\vec{u^{r}}$ $\gets$ $\argmax \limits_{\vec{u}^{k}}$ ($\gamma^{k} \cdot N^{k}_{g}$) \\
    \Return $\vec{u^{r}}, N_{g}^{r}$
    \caption{$\mu$-MOG}\label{alg:grasp-planner}
\end{algorithm}
\setlength{\floatsep}{2mm}
Alg.~\ref{alg:decluttering-algorithm} details the decluttering algorithm, which is similar to the picking algorithm in prior work \cite{Agboh-ISRR-2022}.
We estimate the current state $\vec{x}$, containing $N_{o}$ objects, using color segmentation to isolate objects \textcolor{black}{ and using their convex hulls to find vertices} (line 2). The algorithm then uses the subroutine CreateObjGroups(.) (line 3) to create a set of distinct object groups. It loops through center points of objects and creates groups of all objects that are half a gripper width radius away. This also includes all single object groups. Then, the RankObjGroups(.) subroutine (line 4) ranks the list of object groups
by their size. The RobustGraspPlanner(.) subroutine (line 6) in section \ref{sec:rob-mog-plan} finds a grasp for the largest object group. The grasp is executed and the whole process repeats until the table is cleared or a time limit is reached.

\subsection{Robust multi-object grasp planning}\label{sec:rob-mog-plan}

One main distinction from prior work is the robust multi-object grasp planner $\mu$-MOG, which is detailed in Alg.~\ref{alg:grasp-planner}. The algorithm generates multiple grasp candidates (line 1) using GenGraspCands(.). It finds the convex hull of a given group of objects and generates $N_{p}$ points that uniformly cover the convex hull. At each point, it generates $N_{\theta}$ orientation samples. It rejects grasp samples that result in collisions between the gripper jaws and any object. Next, it loops through grasp candidates (lines 2-4) and estimates (i) the probability $\gamma^{k}$ of satisfying necessary conditions using NecessaryCondsProba($\vec{x}$, $\vec{u}^{k}$), and (ii) the predicted number of objects that $\vec{u}^{k}$ will successfully grasp $N_{g}^{k}$, using \textit{MOG-Net}.

To calculate $\gamma^{k}$, NecessaryCondsProba(.) performs Monte-Carlo sampling so that the relative position of the grasp candidate varies with respect to the position of the objects in the group. Specifically, we consider samples $\vec{u'} = \vec{u} + \delta{\vec{u}}$, and $\vec{x'} = \vec{x} + \delta{\vec{x}}$ where $\delta{\vec{u}} \sim \mathcal{N}(0, \vec{\sigma_{u}}^2)$, $\delta{\vec{x}} \sim \mathcal{N}(0, \vec{\sigma_{x}}^2)$, and $\sigma_{\vec{u}}$, $\sigma_{\vec{x}}$ are standard deviations for control and state respectively. Then, it returns the ratio of grasp samples ($\gamma^{k}$) that satisfy the necessary conditions, under state and control uncertainty. Finally, we choose the robust grasp $\vec{u^{r}}$ such that:
\begin{align}
    \vec{u^{r}} = \argmax \limits_{\vec{u}^{k}} (\gamma^{k} \cdot N_{g}^{k})
\end{align}

\subsection{Robustness to Frictional Uncertainty}
We assumed a lower bound on the coefficient of friction for all objects in Sec.~\ref{sec:assumptions}. This is a conservative assumption to allow for frictional uncertainty. Lower values of $\mu$ mean fewer admissible grasp candidates. Thus, any grasp candidate that satisfies the necessary conditions for the lower bound will also satisfy the conditions for higher values of $\mu$.

\section{Necessary conditions for multi-object grasping}\label{sec:frictional-mog}
Prior work~\cite{Agboh-ISRR-2022} studied frictionless multi-object grasping. In this work, we extend the analysis to include friction and derive the frictional necessary conditions to \textcolor{black}{achieve frictional force closure.}

\subsection{Frictional equilibrium multi-object grasps}
\label{sec:friction-equi-grasps}

Under the frictionless point contact model, the number of possible antipodal grasp configurations for a polygon is limited. For example, a triangle has only one possible equilibrium grasp configuration: a vertex and an opposing edge. With friction, it is possible to achieve more equilibrium grasps, which is dependent on the coefficient of friction. We first analyze equilibrium grasps for a single object under the frictional point contact model and extend those results to multiple objects.

In single object grasping with a parallel-jaw gripper, frictional equilibrium grasps occur at pairs of contacts where the friction cones contain opposing forces that lie on the line passing through them. Recall that $\mu$ is the lower bound on the coefficient of friction at the left and right contacts.
Then, the friction cones ($C_{l}$ and $C_{r}$) are characterized by $\alpha_{l} = \alpha_{r} = \tan^{-1}(\mu)$, and are centered on the contact normals ($\hat{\vec{n}}_{l}$, $\hat{\vec{n}}_{r}$). If the line $L_g$ that passes through both contact points is contained in both friction cones, the parallel-jaw grasp is in equilibrium.

For a convex polygonal object, it is then possible to consider discrete points along the object's surface, and enumerate equilibrium grasps by considering opposing contact pairs using the friction cones. The number of equilibrium grasps  can be infinite depending on the size of the friction cone ($\alpha$).

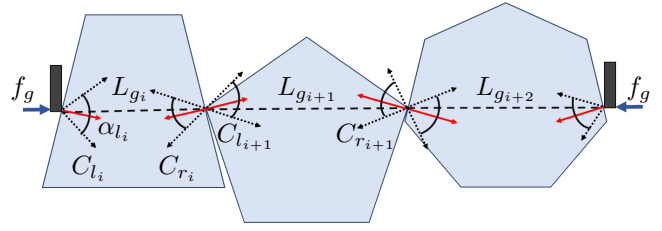
\begin{figure}[t]
    \centering
    \includegraphics[scale=0.35]{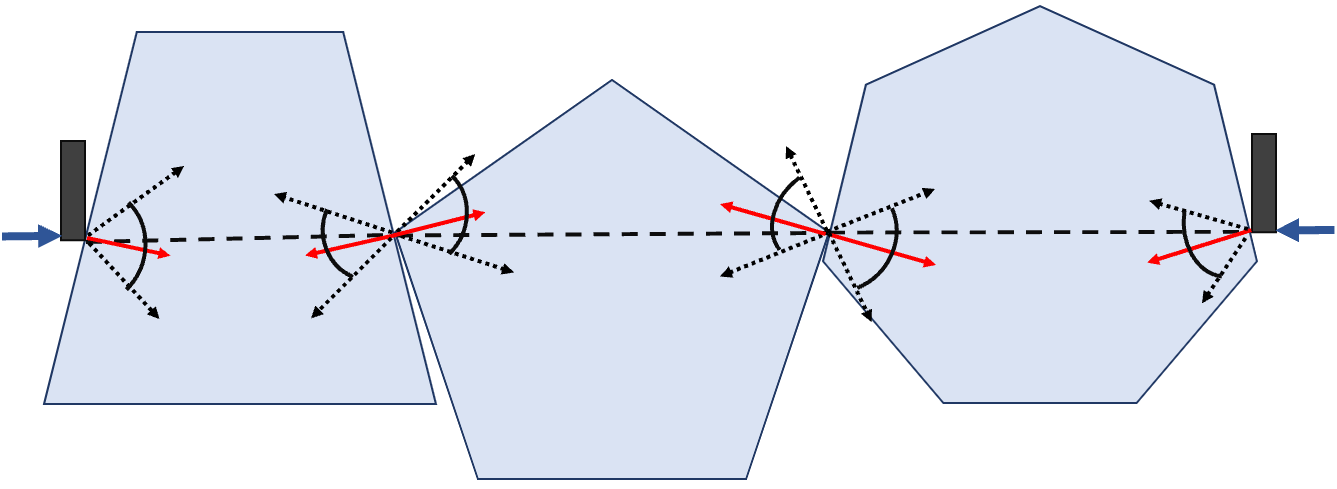}
    \begin{picture}(0,0)
    \put(-243,50) {$f_g$}
    \put(-10,50) {$f_g$}
    \put(-210,35) {$\alpha_{l_i}$}
    \put(-220,20){$C_{l_i}$}
    \put(-187,20){$C_{r_i}$}
    \put(-205,50) {$L_{g_i}$}
    \put(-165,32){$C_{l_{i+1}}$}
    \put(-120,32){$C_{r_{i+1}}$}
    \put(-142,50){$L_{g_{i+1}}$}
    \put(-67,50) {$L_{g_{i+2}}$}
    \end{picture}
    \caption{A top-down view of a 3-object frictional equilibrium grasp. We check that each object $i$ is in an equilibrium grasp by inspecting their left ($C_{l_i}$) and right ($C_{r_i}$) friction cones and the line (${L}_{g_i}$) passing through both contact locations. Also, we ensure that all connecting lines lie on the same line ($\hat{\vec{L}}_{g_{i}} = \hat{\vec{L}}_{g_{i+1}} = \dots = \hat{\vec{L}}_{g_{n_{o}}}$). The friction cones are centered on the contact normal and are defined by the coefficient of friction $\mu$. $\alpha = tan^{-1}(\mu)$.}
    \label{fig:frictional-multi-object}
\end{figure}

We require each individual object in a frictional multi-object grasp to be in an equilibrium grasp. Consider Fig.~\ref{fig:frictional-multi-object}. It shows a sample frictional multi-object equilibrium grasp for 3 objects. The left ($C_{l_i}$) and right ($C_{r_i}$) friction cones for object $i$ are characterized by $\alpha_{l_i} = \alpha_{r_i} = \tan^{-1}(\mu)$. To achieve an equilibrium multi-object grasp for a group of $n_{o}$ objects, both friction cones for each object $i \in \{0,1,\dots, n_{o}-1\}$, must contain opposing forces that lie on line $L_{g_i}$, connecting their contact locations. Since the forces at object-object contacts are reactionary, all connecting lines ($L_{g_i}$ in Fig. \ref{fig:frictional-multi-object}) must lie on the same line:
\begin{align}\label{eq:connecting_line}
    \hat{\vec{L}}_{g_{i}} = \hat{\vec{L}}_{g_{i+1}} = \dots = \hat{\vec{L}}_{g_{n_{o}}}
\end{align}
where $\hat{\vec{L}}_{g_{i}}$ is the unit vector of $L_{g_{i}}$, for object $i$.

\subsection{Frictional multi-object grasp diameter}
\label{sec:friction-grasp-diameter}
Every object has a \textit{final} diameter $d_f$ at which a stable
frictional single-object grasp will occur. This is the distance
between gripper jaws when they become stationary in a stable
grasp. By sampling stable frictional contact pairs on an object,
we can enumerate multiple final single object grasp diameters
to find the minimum, $d^{*}_{f}$. Similarly, every object group has
a \textit{final} multi-object diameter $h_{f}$ at which a stable frictional
multi-object grasp will occur. We compute the minimum \textit{final} multi-object grasp diameter $h^{*}_{f}$ given $n_{o}$ objects as:
\begin{align} \label{eq:multi-object-diameter-comp}
    h^{*}_{f} = \sum^{n_{o}-1}_{i=0} d^{*}_{f_{i}}
\end{align}
This is true because we showed in Eq. ~\ref{eq:connecting_line} that all connecting
lines must lie on the same line. We use $h^{*}_{f}$ in the multi-object
grasping necessary condition detailed in Sec. ~\ref{sec:necc_conds}.

\subsection{Necessary conditions for multi-object grasping}
\label{sec:necc_conds}
Prior work on frictionless multi-object grasping \cite{Agboh-ISRR-2022} developed two necessary conditions --- intersection area and multi-object grasp diameter. These conditions are used to filter inadmissible grasps in a multi-object grasp planner. In this section we summarize these conditions for completeness.

Given a grasp, let the internal rectangular region between the gripper jaws be S. Let $o^{s}_{i} = S \cap \vec{o}_{i}$, be the intersection polygon between $S$ and object $\vec{o}_i$.

\subsubsection{Intersection area}
\label{sec:intersection-area}
Let $A_{i}(t) = \mathrm{Area}(o^{s}_{i})$, be the area of the intersection polygon for object $i$, during a multi-object grasp at time $t$ for a grasp. The intersection area condition from prior work \cite{Agboh-ISRR-2022} can be written as:
\begin{align}\label{eq:relaxed_int_area}
 A_{i} (0) > 0, \hspace{1mm} i \in \{0,1,\dots, n_{o}-1\}.
\end{align}
We directly use this intersection area condition in this work.
\subsubsection{Multi-object grasp diameter}
Prior work~\cite{Agboh-ISRR-2022} defined the multi-object grasp diameter necessary condition. We restate it here for completeness. Let $w_g (t)$ be the gripper width at time $t$. Let $b_{l}(t)$ be the shortest distance between $o^{s}_{0}$ and the left jaw (where $o^{s}_{0}$ is the closest object to the left jaw), and $b_{r}(t)$ be the shortest distance between $o^{s}_{n_{o}-1}$ and the right jaw (where $o^{s}_{n_{o}-1}$ is the closest object to the right jaw). Then, the multi-object grasp diameter as a function of time is: $h(t) = w_{g}(t) - (b_{l}(t) + b_{r}(t))$. Let $h_{0}$ be the \textit{initial} multi-object grasp diameter at time $t_{0}$, and $h_{f}$ be the corresponding \textit{final} multi-object grasp diameter, at time $t_{f}$ when the grippers become stationary after closing.
Given a group of $n_o$ objects, one can compute the minimum possible diameter $h^{*}_{f}$, such that any multi-object grasp must satisfy:
\begin{align}\label{eq:multi-object-diameter}
    h_{0} \geq h^{*}_{f}.
\end{align}

Prior work \cite{Agboh-ISRR-2022} computed $h^{*}_{f}$ for the frictionless case. Here we provide a method to compute it for the frictional case. Specifically, we compute $d^{*}_{f_{i}}$ in Eq.~\ref{eq:multi-object-diameter-comp} by sampling $N_{s}$ contact points along an object's edge. Then, we generate all contact pairs resulting from these points. Next, we check if a contact pair is stable by ensuring that the left and right friction cones contain the line connecting the contact points. We pick $d^{*}_{f_{i}}$ as the stable contact pair with the minimum diameter.

Note that given a group of $n_{o}$ objects, $h^{*}_{f}$ is smaller in the frictional case compared to the frictionless case. This allows for more multi-object grasps to satisfy the necessary conditions when friction is considered. \textcolor{black}{We further note that these necessary conditions are independent of the contant area, given the Coulomb friction assumption.}

\section{Learning a multi-object grasp neural network}
\label{sec:mog-net}

We train \textit{MOG-Net} with self-supervised learning in real to predict the number of objects ($N_{g}$) that can be successfully grasped from a target object group. It takes the state of all objects in a target group, and a grasp action $\vec{u}$ as inputs. In Sec.~\ref{sec:data_collection} we detail our data collection process, and in Sec.~\ref{sec:model_structure} we explain details of the neural network model.

\subsection{Data collection}
\label{sec:data_collection}

Physical robot time is expensive and we would like to quickly generate a high quality dataset for \textit{MOG-Net}. An important question is what grasps do we execute during data collection. Given an object group, instead of randomly sampling grasps, we propose to use the frictional multi-object grasp necessary conditions to filter out inadmissible grasps.

Our data collection algorithm is similar to the decluttering algorithm with two key differences: i) unlike in Alg.~\ref{alg:decluttering-algorithm}, during data collection, $obj\_group$ consists of only multi-object groups (i.e $len(obj\_group)>2$), and is chosen at random, ii) in Alg.~\ref{alg:grasp-planner},
we use a heuristic instead of \textit{MOG-Net} --- the total intersection area, $A_{T}$ = $\sum^{n_{o}-1}_{i=0} (A_{i})$ (see Sec.~\ref{sec:intersection-area}). We pick the grasp with $\argmax \limits_{u^{k}}(\gamma^{k} \cdot A_{T}^{k})$. After grasp execution, the data collection system uses an overhead image of the scene and the gripper jaw position to count the number of objects grasped. In this way, data collection is self-supervised.

\subsection{Multi-object grasp neural network}
\label{sec:model_structure}
\textit{MOG-Net} predicts $N_{g}  \in \{0, \dots, N^{max}_{g} \}$, where $N^{max}_{g}$ is the maximum number of objects that can be grasped. We train a separate classifier for each $N_{g}$ prediction class, using the same dataset collected in real. Specifically, data for a specific class is created by setting only occurrences of the desired $N_{g}$ label to true while others are false. Thereafter, we train a feedforward neural network model for each class to perform binary classification. At test time, given a target object group of size $n_{o} \leq N^{max}_{g}$, we query neural network models for classes between $0$ and $n_{o}$. Then, we pick the prediction with the maximum probability as $N_{g}$. We provide further details on \textit{MOG-Net} in \ref{sec:exp_details}.
\section{Physical Experiments}
\label{experiments}
\begin{center}
\begin{table}[b]
\caption{In this table, we show the number of grasped objects in the dataset of 1545 grasp samples collected on the physical robot for \textit{MOG-Net} and \textit{Rand-Net}. We see that \textit{MOG-Net} produces a more balanced and higher quality dataset compared to \textit{Rand-Net}. This saves valuable robot time. }
\centering
\begin{tabular}{@{}l*{5}{p{3em}}}
\toprule
 & \multicolumn{5}{c}{\hspace{-7mm} Number of grasped objects in the dataset \vspace{1mm}
} \\
\quad & 0 & 1 & 2 & 3 & 4 \\
\midrule
Rand-Net & 543 & 665  & 250 & 74 & 13\\
MOG-Net & 278 & 322 & 474 & 341 & 130\\
\bottomrule
\end{tabular}
\label{table:data_collection}
\end{table}
\end{center}
\begin{center}
\begin{table*}[t]
\caption{Physical decluttering experimental results for 10 scenes, each with 58 objects randomized as described in \ref{sec:exp_setup}. We reset each scene precisely by hand to compare the methods. Errors here are within 95\% confidence interval of the mean. Compared to prior work (Frictionless MOG\cite{Agboh-ISRR-2022}), \textit{MOG-Net} achieved \textbf{13.7\%} higher grasp success, \textbf{1.6x} PPH, and plans grasps \textbf{6.3x }faster. We also record a \textbf{3.1x} improvement in PPH compared to Frictional SOG.}
\centering
\begin{tabular}{@{}l@{\quad}c@{\quad}c@{\quad}c@{\quad}c@{\quad} c@{\quad}c@{}}
\toprule
Methods & Success rate (\%) & Picks per hour  & Grasped Objs. & Planning time (s) & Cleared (\%) & Pick attempts\\
\midrule
Frictional SOG & 51.5 $\pm$ 2.2 & 188.9 $\pm$ 9.2 & 0.59 $\pm$ 0.0
& \textbf{0.11 $\pm$ 0.0} & 98.5 $\pm$ 0.7 & 80.7 $\pm$ 3.6\\
Rand-Net & 67.0 $\pm$ 3.0 & 321.9 $\pm$ 19.9 & 0.93 $\pm$ 0.1 &
 0.53 $\pm$ 0.1 & 99.5 $\pm$ 0.7 & 54.9 $\pm$ 3.3\\
Frictionless MOG \cite{Agboh-ISRR-2022}  & 68.9 $\pm$ 5.6 & 370.5 $\pm$ 48.3 & 1.30 $\pm$ 0.2 & 2.21 $\pm$ 0.2  & 99.0 $\pm$ 0.8 & 43.0 $\pm$ 5.6 \\
Frictionless MOG-Net & 71.5 $\pm$ 5.7 & 392.1 $\pm$ 24.8 & 1.26 $\pm$ 0.1 & 0.59 $\pm$ 0.1 & 97.8 $\pm$ 1.4 & 44.0 $\pm$ 2.4 \\
MOG-Net & \textbf{82.6 $\pm$ 3.8} & \textbf{580.7 $\pm$ 29.1} & \textbf{1.83 $\pm$ 0.1} & 0.35 $\pm$ 0.1 & \hspace{-2.2mm} \textbf{100.0 $\pm$ 0.0} & \textbf{30.6 $\pm$ 1.6}\\
\bottomrule
\end{tabular}
\label{tab:decluttering_results}
\end{table*}
\vspace{-15mm}
\end{center}
We conduct physical experiments to evaluate the data collection and decluttering algorithms. Our goal is to investigate the effect of friction on multi-object grasping for different methods. In the following subsections, we explain the general setup, experimental details, baselines, and results.

\subsection{Experimental setup}\label{sec:exp_setup}
The setup is as shown in Fig.~\ref{fig:fig1} where we use a UR5 robot with a Robotiq 2F-85 gripper. In experiments, we have two sets of objects -- low friction and frictional. Each contains a total of 58 objects from 3-sided to 8-sided convex polygons. To get frictional objects we wrap low friction objects with transparent, \textit{non-stick}, high friction tape.

During data collection and decluttering experiments, we generate initial scenes with randomized object poses. We begin by repeatedly creating random object clusters. Each scene contains 17 non-overlapping object clusters that have a random center point. Within each cluster, we randomly sample \footnote{For each cluster, we first randomly select an orientation for the diameter of that cluster and an ordered subset of all 58 objects without replacement to place in that cluster. We then randomly sample points along the diameter of that cluster, with uniform noise $[-0.9, 0.9]$ cm perpendicular to that diameter, to be the center of the longest edge for that object. For each object in the cluster, we sample a random orientation in $[-\pi/2, \pi/2]$.} the number of objects, their types, positions, and orientations.

We use an RGBD camera (Intel Realsense Camera D435) to get a top-down image of the cluttered scene and then extract vertices of all objects to get the state $\vec{x}$. The grasp action $\vec{u}$ involves four steps. (1) Moving the open gripper above the desired grasp pose and lowering until just above the table. (2) Closing the gripper jaws. (3) Moving the gripper upwards and above the packing box. (4) Opening the jaws so the objects fall into the packing box. All parameters used in this work are detailed here (i) grasp sampling parameters: $N_{p}=25$ , and $N_{\theta}=12$. (ii) Monte-Carlo sampling parameters to estimate $\gamma$: $\vec{\sigma_{u}}=[2mm,~2mm,~2^\circ]$, and $\vec{\sigma_{x}} = 2 \cdot \{\vec{1}\}mm$. (iii) friction: $\mu = 0.5$ for frictional, $\mu = 0.01$ for frictionless, and $N_{s}=5$ for contact point sampling. (iv) $N_{v}$, the maximum number of vertices per object is set to 8 in our object set.

\subsection{Baseline methods}

\subsubsection{Frictional SOG} We use Alg.~\ref{alg:decluttering-algorithm} but restrict object groups to contain only single objects. It plans a frictional single-object grasp with the frictional point-contact model.
\textcolor{black}{This is similar to state of the art single-object grasping methods such as Dex-Net}.

\subsubsection{Rand-Net} This trains the same neural network model as ours but with a different dataset. The dataset is comprised of random grasps to execute during training from $grasp\_cands$ in Alg.~\ref{alg:grasp-planner} but without filtering with the necessary conditions. We trained and tested this baseline with frictional objects.

\subsubsection{Frictionless MOG}
This is state-of-the-art in multi-object grasping from prior work \cite{Agboh-ISRR-2022}. It filters grasp candidates with frictionless necessary conditions and uses a physics simulator (Mujoco) to find grasps. 

\subsubsection{Frictionless MOG-Net} This baseline uses \textit{MOG-Net} but computes necessary conditions using a low friction value. 

\textcolor{black}{Note that we use low-friction objects for the two frictionless baselines above since they rely on frictionless necessary conditions from prior work \cite{Agboh-ISRR-2022} to generate grasps.}

We left one set unmodified to use as low-friction objects for the frictionless baselines. We added clear grip tape to the objects of the other set to use as the high-friction counterparts for the frictional methods. The low-friction objects do not generate frictional grasps, but instead get filtered out as inadmissible grasps using the frictionless necessary conditions from prior work \cite{Agboh-ISRR-2022}. 

\subsection{Experimental details}
\label{sec:exp_details}
\subsubsection{Data collection}
We collected 1545 grasp samples in real for \textit{MOG-Net} and \textit{Rand-Net}, using the random initialization process described in ~\ref{sec:exp_setup}. 
\subsubsection{Decluttering}
We create 10 decluttering scenes with the same process as data collection. In each scene, we use the four different baselines and \textit{MOG-Net} to generate grasps. We replicated the randomly generated scene manually in each case, leading to a total of 50 physical robot experiment scenes and a total of 2532 robot grasp samples. A failed grasp attempt is where
the robot misses a grasp (all objects escape), or where all objects fall out of the gripper
before they reach the packing box.

\subsubsection{Neural network details} We use a feedforward neural network with 4 hidden layers. Specifically, we use the MLPClassifier(.) from scikit-learn \cite{Pedregosa-JMLR-2011} with default parameters and $hidden\_layer\_sizes=(500, 300, 150, 50)$. We limit the maximum number of objects for the input vector to $N^{max}_{g}=4$ given the gripper's size. Each object can have at most ($N_{v}=8$) vertices. The input vector contains x and y points for each object's vertex taken with respect to the grasp center. It also contains the grasp orientation. Therefore, the input vector size is fixed at 16 $\times$ 4 + 1 = 65. If an object has vertices less than $N_{v}$, we pad the input vector with the last vertex in the list. Similarly if the number of objects in a group ($n_{o}$) is less than $N^{max}_{g}$, we pad the input vector with the last object's vertices in the list. Recall that we train 5 classifiers with the same model architecture to predict 5 different \textit{number of objects grasped ($N_{g}$)} classes (0 to 4). We then pick the $N_{g}$ prediction with the highest probability as the output of \textit{MOG-Net}.

\subsection{Metrics}
We compare methods with (i)\textit{ Success rate}: percentage of grasp attempts that moved at least one object into the box, (ii) \textit{Picks per hour}: total number of objects picked per hour, through single or multi-object grasps, (iii) \textit{Grasped Objs}: the average number of objects grasped per pick attempt, (iv) \textit{Planning time}: time to plan a grasp, (v) \textit{Cleared}: fraction of objects that were moved to the box from the cluttered scene, (vi) \textit{Pick attempts}: the average number of pick attempts.
\subsection{Results}
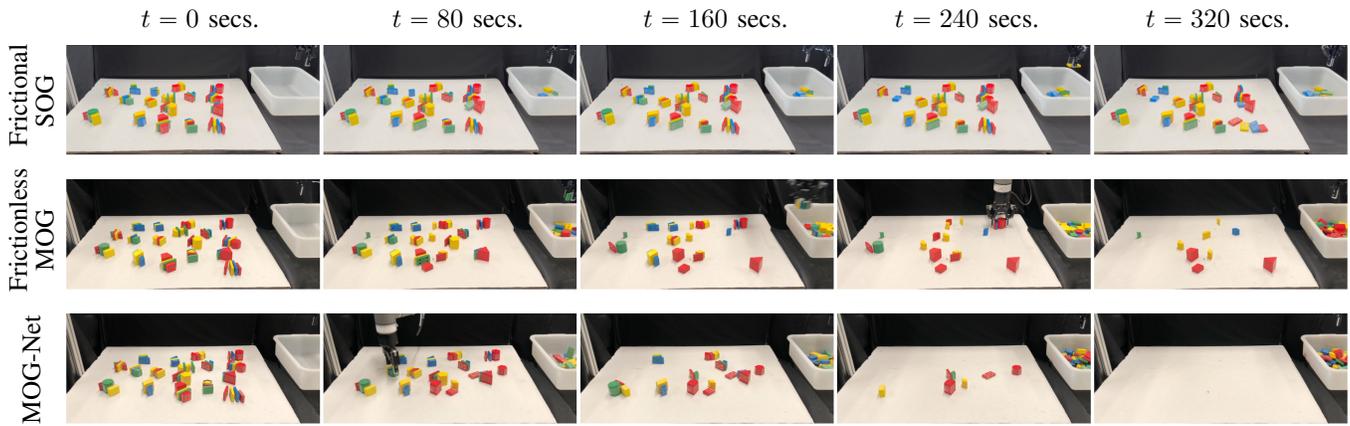
\begin{figure*}[htb!]
    \vspace{6mm}
    \hspace{8mm}
\includegraphics[scale=0.16]{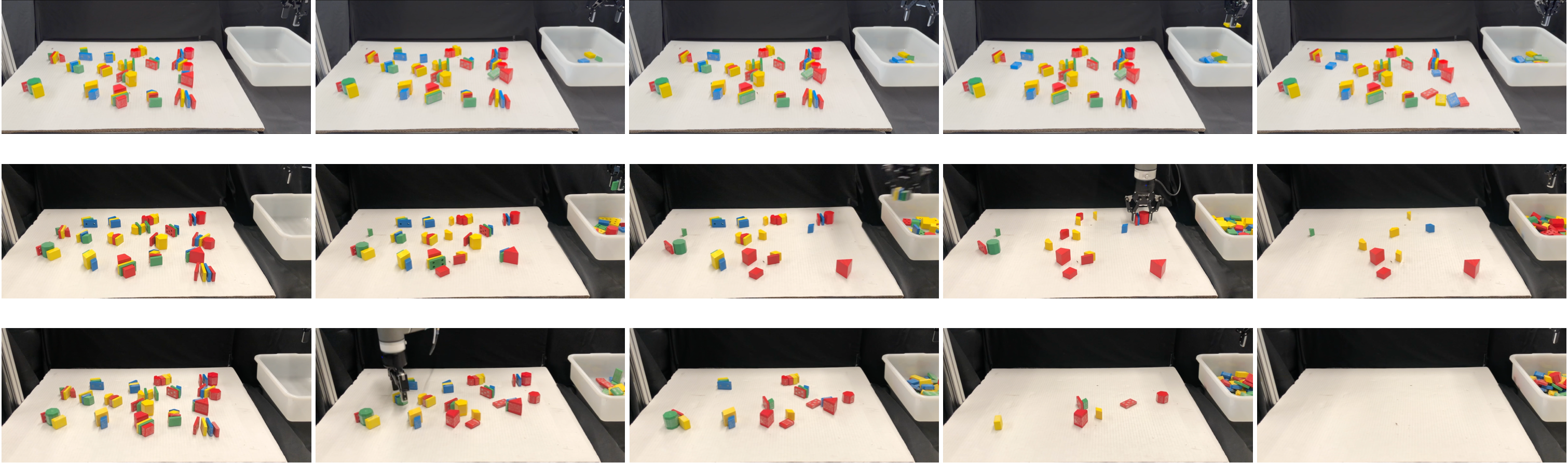}
        \begin{picture}(0,0)
        \put(-460, 150) {$t=0$ secs.}
        \put(-365, 150) {$t=80$ secs.}
        \put(-270, 150) {$t=160$ secs.}
        \put(-175, 150) {$t=240$ secs.}
        \put(-80, 150) {$t=320$ secs.}
        \put(-510, 105) {\rotatebox{90}{Frictional}}
        \put(-500, 112) {\rotatebox{90}{SOG}}
        \put(-510, 50) {\rotatebox{90}{Frictionless}}
        \put(-500, 58) {\rotatebox{90}{MOG}}
        \put(-505, 0) {\rotatebox{90}{MOG-Net}}
    \end{picture}
    \caption{The same decluttering scene, replicated across three methods. \textit{Frictional SOG} is the single object grasping baseline, \textit{Frictionless MOG} is prior work \cite{Agboh-ISRR-2022}, and \textit{MOG-Net} is the method proposed in this paper. \textit{MOG-Net} successfully removed all objects in 320 seconds while the other methods were slower.}
    \vspace{-5mm}
         \label{fig:scene_setups}
\end{figure*}

\subsubsection{Data collection and model}
Please see Table ~\ref{table:data_collection}
for a distribution of the data collected for \textit{MOG-Net} and \textit{Rand-Net}. We see that \textit{MOG-Net} collects a well distributed dataset with more balanced samples per class as opposed to \textit{Rand-Net}, thereby saving valuable physical robot time. Classification accuracy was 71.8\% for \textit{MOG-Net} and 45.1\% for \textit{Rand-Net} on a combined test set containing 20\% of the data samples.

\subsubsection{Decluttering}
The results can be found in Table ~\ref{tab:decluttering_results}. We see that introducing friction and learning a grasp function in real significantly increases PPH. Compared to prior work \cite{Agboh-ISRR-2022}, \textit{MOG-Net} achieved 13.7\% higher grasp success, 1.6x picks per hour, and plans grasps 6.3x faster. We also record a 3.1x improvement in PPH compared to single object grasping (Frictional SOG, which is very similar to other state of the art grasping methods such as Dex-Net \cite{mahler2016dex}) as opposed to a 1.6x improvement in prior work \cite{Agboh-ISRR-2022}. A sample rollout on the same scene for 3 of the methods can be seen in Fig. \ref{fig:scene_setups}.

\textit{MOG-Net} outperforms \textit{Rand-Net} on all metrics, suggesting the importance of our data collection system that generates a more balanced dataset for training by using the necessary conditions.
Frictionless MOG uses a physics simulator and that led to a lower success rate, compared to \textit{MOG-Net} which was trained in real.
Frictionless \textit{MOG-Net} outperforms Frictionless MOG on PPH. This demonstrates the importance of a reduced planning time by using a learned model, instead of a physics simulator.
One mode of failure for all systems is where a grasp attempt topples objects, resulting in difficult-to-grasp poses.
\vspace{-0.5mm}
\section{Limitations and Future Work}
This work has the following limitations: 
\noindent
i) \textit{Pushing to rearrange objects}:
\textcolor{black}{The experiments have randomly generated object groups, but the robot could consider pushes to rearrange objects before planning multi-object grasps. This is explored in \cite{aeron2023pushmog, srinivas2023busboy}.}
\noindent
ii) \textit{Contact models}:
 To derive the necessary conditions to speed up MOG-Net's training and grasp planning, we assumed a frictional point-contact model. However, for more general 3D objects we will explore soft contact models that account for torsional frictional forces. \textcolor{black}{iii) \textit{Non-polygonal objects}:
 We assume scenes with extruded convex polygons but household objects can be curved, non-convex, and non-polygonal. We will explore multi-object grasps for household objects and in more diverse backgrounds.}

In this work, we consider the decluttering problem where multiple convex polygonal objects are grasped and moved to a packing box. We leverage a novel frictional multi-object grasping necessary condition to train \textit{MOG-Net}, a neural network model using real examples. It predicts the number of objects grasped out of a target object group. We use \textit{MOG-Net} in a novel grasp planner to generate robust multi-object grasps. Experiments suggest that introducing friction and considering robustness in multi-object grasping leads to improvements in success rate and picks per hour, compared to prior work.

\section{Acknowledgement}
\small{
This research was performed at the AUTOLAB at UC Berkeley in affiliation with the Berkeley AI Research (BAIR) Lab, and the CITRIS “People and Robots” (CPAR) Initiative. The authors were supported in part by donations from Siemens, Toyota Research Institute, Bosch, Google, and Autodesk and by equipment grants from PhotoNeo, NVidia, and Intuitive Surgical. Mehmet Dogar was partially supported by an EPSRC Fellowship (EP/V052659). For the purpose of open access, the authors have applied a Creative Commons Attribution (CC BY) license to any Accepted Manuscript version arising.}
\bibliographystyle{unsrtnat}
{\let\clearpage\relax \vspace{0 mm} \bibliography{main} }
\newpage

\end{document}